\documentclass[10pt, conference, compsocconf]{IEEEtran}
\usepackage{cite}

\ifCLASSINFOpdf
  \usepackage[pdftex]{graphicx}
\else
  \usepackage[dvips]{graphicx}
\fi
\usepackage[cmex10]{amsmath}
\usepackage{algorithmic}

\usepackage{array}
\usepackage{amssymb}
\usepackage{flushend}
\usepackage{bm}     
\usepackage{url}

\usepackage{subcaption} 
\DeclareCaptionFont{footnotesize}{\footnotesize} 
\usepackage{arabtex}

\usepackage{siunitx}

\graphicspath{{./Figs/}}

\begin{document}
%
\title{Adversarial Generation of Handwritten Text Images Conditioned on Sequences}


\author{\IEEEauthorblockN{Eloi Alonso
\IEEEauthorblockA{A2iA SA, Paris, France}
\IEEEauthorblockA{\'{E}cole des Ponts ParisTech}
}
\and
\IEEEauthorblockN{Bastien Moysset
\IEEEauthorblockA{A2iA SA, Paris, France}
}
\and
\IEEEauthorblockN{Ronaldo Messina
\IEEEauthorblockA{A2iA SA, Paris, France}
}
}



%


\maketitle

\begin{abstract}
State-of-the-art offline handwriting text recognition systems tend to use neural networks and therefore require a large amount of annotated data to be trained. In order to partially satisfy this requirement, we propose a system based on Generative Adversarial Networks (GAN) to produce synthetic images of handwritten words. We use bidirectional LSTM recurrent layers to get an embedding of the word to be rendered, and we feed it to the generator network. We also modify the standard GAN by adding an auxiliary network for text recognition. The system is then trained with a balanced combination of an adversarial loss and a CTC loss. Together, these extensions to GAN enable to control the textual content of the generated word images. We obtain realistic images on both French and Arabic datasets, and we show that integrating these synthetic images into the existing training data of a text recognition system can slightly enhance its performance.
\end{abstract}

%

%
\IEEEpeerreviewmaketitle


\section{Introduction}

Having computers able to recognize text from images is an old problem that has many practical applications, such as automatic content search on scanned documents. Transcribing \textit{printed} text is now a reliable technology. However, automatically recognizing \textit{handwritten} text is still a hard and open problem. Unlike printed text, cursive handwriting cannot be segmented into individual characters, since their boundaries are ill-defined. Graves et al. \cite{Graves:06icml} introduced the Connectionist Temporal Classification (CTC), a loss enabling to train neural networks to recognize sequences without explicit segmentation. Today, in order to deal with such a complex problem, state-of-the-art solutions \cite{Bluche2017,8270157} are all based on deep neural networks and the CTC loss. 

The supervised training of these neural networks requires large amounts of annotated data; in our case, images of handwritten text with corresponding transcripts. However, annotating images of text is a costly, time-consuming task. We therefore propose a system to reverse the annotation process: starting from a given word, we generate a corresponding image of cursive text. We first tackle the challenge of generating realistic data, and then address the question of using such synthetic data to train neural networks in order to improve the performance of handwritten text recognition.

\begin{figure}[ht]
    \centering
    \includegraphics[width=0.9\columnwidth]{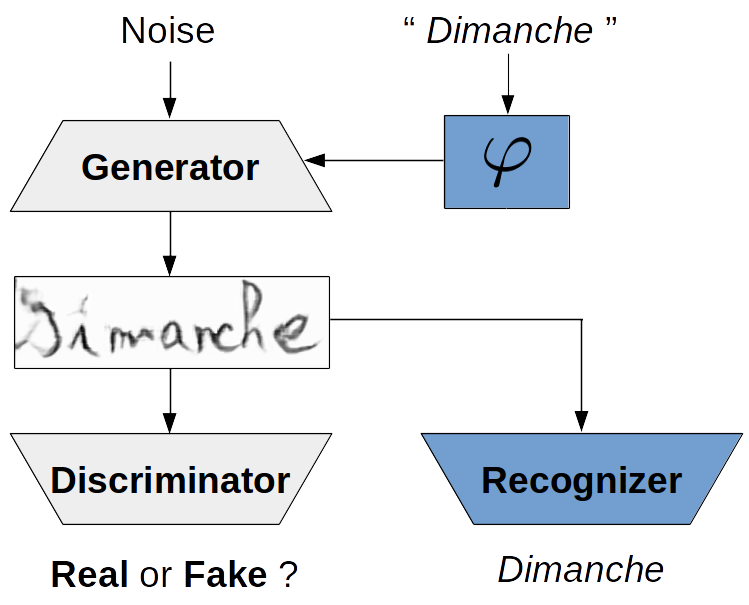}
    \caption{Adversarial generation of an image of text, conditioned on the textual content (``Dimanche''). The differences with a standard GAN are shown in blue.}
    \label{fig:1_principle}
\end{figure}

The problem of generating images of handwritten text has already been addressed in the past. Many techniques \cite{Elanwar13} are based on a collection of templates of a few characters, either human-written or built using Bezier curves. These templates are possibly perturbed and finally concatenated. However, this class of solutions, that simply concatenates character models, cannot faithfully reproduce the distribution of real-world images. It is also complex to have templates that are generic enough to result in truly cursive text. Alternatively, following the online approach, we can consider the handwriting as a trajectory, typically recorded with pen computing. In this setting, the model aims at producing a sequence of pen positions to generate a given word. Graves et al. \cite{Graves13} use a Long Short-Term Memory (LSTM) \cite{Hochreiter,Gers2000} recurrent neural network to predict such a sequence, and let the network condition its prediction on the target string to synthesize handwriting. However, this method does not allow to deal with some features useful for offline recognition, such as background texture or line thickness variations.

Generative Adversarial Networks (GAN) \cite{NIPS2014_5423} offer a powerful framework for generative modeling. This architecture enables the generation of highly realistic and diverse images. 
The original GAN does not allow any control over the generated images, but many works \cite{Mirza2014,Odena2016,Brock18} proposed a modified GAN for class-conditional image generation. However, we want to condition our generation on the sequence of characters to render, not on a single class. Closer to our goal, Reed et al. \cite{reed2016} conditions the generation on a textual description of the image to be produced. In addition to a random vector, their generator receives an embedding of the description text, and their discriminator is trained to classify as fake a real image with a non-matching description, to enforce the generator to produce description-matching images.

To the best of our knowledge, there is only one work \cite{chang2018generating} on a GAN for text image synthesis. While our generation process is directly conditioned on a sequence of characters, this method follows a style transfer approach, resorting to a CycleGAN \cite{Zhu17} to render images of isolated handwritten Chinese characters from a printed font.

Data augmentation techniques based on distortion and additive noise do not allow to enlarge the textual contents of the training data. Moreover, having control of the generated text enables the creation of training material that covers even rare sequences of characters, which can be expected to improve the recognition performance. This, combined with the intrinsic diversity, provides a strong motivation to use a conditional GAN for the generation of cursive text images.

In this paper, we make the following contributions:
\begin{itemize}
    \item We propose an adversarial architecture, schematically represented in Fig. \ref{fig:1_principle}, to generate realistic images of handwritten words.
    \begin{itemize}
        \item We use bidirectional LSTM recurrent layers to encode the sequence of characters to be produced.
        \item We introduce an auxiliary network for text recognition, in order to control the textual content of the generated images.
    \end{itemize}
    \item We obtain realistic images on both French and Arabic datasets. 
    \item Finally, we slightly improve text recognition performance on the RIMES dataset \cite{grosicki2009icdar}, using a neural network trained on a dataset extended with synthetic images.
\end{itemize}


\section{Proposed adversarial model}
\label{sec:model}

We introduce here our adversarial model for handwritten word generation. Section \ref{subsec:argan} gives the general idea and defines the training objectives of the different parts. We detail the network architectures in Section \ref{subsec:archi} and describe our optimization settings in Section \ref{subsec:optim}.

\subsection{Auxiliary Recognizer Generative Adversarial Networks}
\label{subsec:argan}

A standard GAN \cite{NIPS2014_5423} comprises a generator ($G$) and a discriminator ($D$) network, shown in gray in Fig. \ref{fig:1_principle}. $G$ maps a random noise $\boldsymbol{z}$ to a sample in the image space. $D$ is trained to discriminate between real and generated (fake) images. Adversarially, $G$ is trained to produce images that $D$ fails to discriminate correctly. These networks hence have competing objectives. 

\begin{center}
    \centering
    \includegraphics[width=0.9\columnwidth]{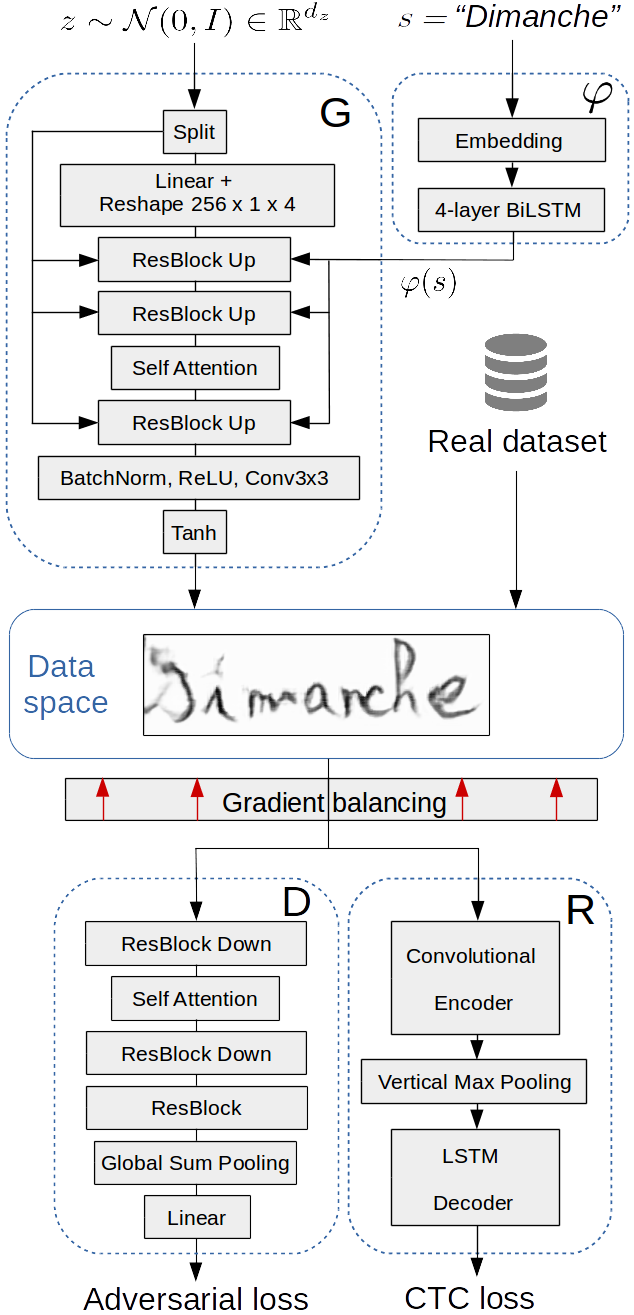}
    \captionof{figure}{Architecture of the networks $\varphi$, $G$, $D$ and $R$. For ease of reading, not all layers are represented (refer to the text for exact details). $G$ receives a chunk of noise and $\varphi(\boldsymbol{s})$ in each ResBlock (details in Fig. \ref{fig:2_resblock}). Both $G$ and $D$ include a self-attention layer. $R$ follows the architecture of \cite{Bluche2017} and is trained with only real data using the CTC loss. We resort to the hinge version of the adversarial loss for $D$. When training $G$, we balance the gradients coming from $D$ and $R$ (details in Section \ref{subsec:optim}).}
    \label{fig:2_model}
\end{center}

In order to control the textual content of the generated images, we modify the standard GAN as follows. First, we use a recurrent network ($\varphi$) to encode $\boldsymbol{s}$, the sequence of characters to be rendered in an image. $G$ takes this embedding $\varphi(\boldsymbol{s})$ as a second input. Then, in the vein of \cite{Odena2016}, the generator is asked to carry out a secondary task. To this end, we introduce an auxiliary network for text recognition ($R$). We then train $G$ to produce images that $R$ is able to recognize correctly, thereby completing its original adversarial objective with a ``collaboration'' constraint with $R$. We use the hinge version of the adversarial loss \cite{Lim2017} and the CTC loss \cite{Graves:06icml} to train this system. Formally, $D$, $R$, $G$ and $\varphi$ are trained to minimize the following objectives:


\begin{align*}
L_D = & - \mathbb{E}_{(\boldsymbol{x}, \boldsymbol{s}) \sim p_{data}} \Big[ \min\big(0, -1 + D(\boldsymbol{x}) \big) \Big] \\
      & - \mathbb{E}_{\boldsymbol{z} \sim p_z, \boldsymbol{s} \sim p_{w}} \Big[ \min\Big(0, - 1 - D\big(G(\boldsymbol{z}, \varphi(\boldsymbol{s}))\big) \Big) \Big]  \\
L_R = & + \mathbb{E}_{(\boldsymbol{x}, \boldsymbol{s}) \sim p_{data}} \Big[ \mathrm{CTC}(\boldsymbol{s}, R(\boldsymbol{x})) \Big] \\
L_{(G, \varphi)} = & - \mathbb{E}_{\boldsymbol{z} \sim p_z, \boldsymbol{s} \sim p_{w}} \Big[ D\big( G(\boldsymbol{z}, \varphi(\boldsymbol{s})) \big) \Big] \\
      & + \mathbb{E}_{\boldsymbol{z} \sim p_z, \boldsymbol{s} \sim p_{w}} \Big[ \mathrm{CTC}\Big(\boldsymbol{s}, R\big( G(\boldsymbol{z}, \varphi(\boldsymbol{s})) \big) \Big) \Big] 
\end{align*}

with $p_{data}$ the joint distribution of real \texttt{[image, word]} pairs, $p_{z}$ a prior distribution on input noise and $p_{w}$ a prior distribution of words, potentially different from the word distribution of the real dataset.

\subsection{Networks architecture}
\label{subsec:archi}

\begin{figure}[ht]
    \centering
    \includegraphics[width=\columnwidth]{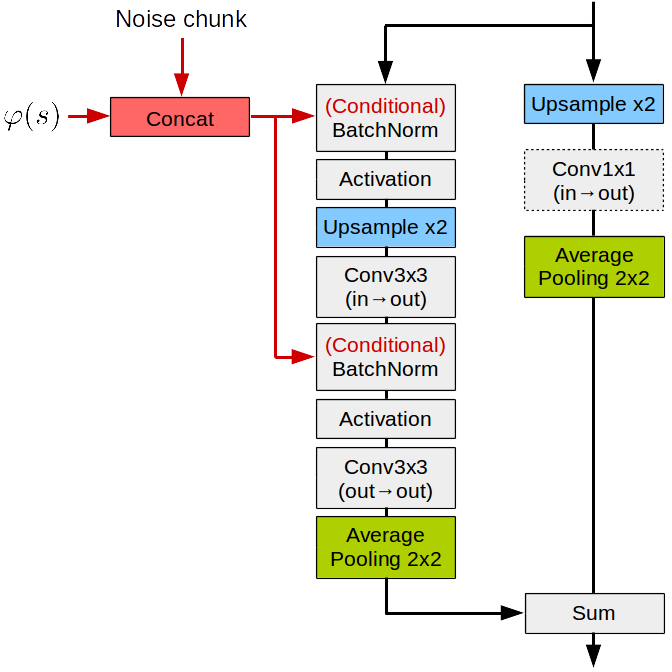}
    \caption{Detail of a ResBlock. The base components are shown in gray. In the ResBlocks of $G$, we concatenate a noise chunk with $\varphi(\boldsymbol{s})$ and feed it to CBN \cite{de2017modulating} layers (red). The unique hidden layer in CBN has 512 units. We also perform up-sampling (blue) with nearest neighbor interpolation. The ResBlocks of $D$ resort to standard Batch Normalization \cite{ioffe} and operate down-sampling (green) with an average pooling. The activation is ReLU \cite{NairH10} in $G$ and LeakyReLU \cite{maas2013rectifier} in $D$. The 1x1 convolution is only used when input (in) and output (out) numbers of channels are different. In the two 3x3 convolutions, the padding and stride are set to 1.}
    \label{fig:2_resblock}
\end{figure}

    
Fig. \ref{fig:2_model} and the text below describe the architecture of networks $\varphi$, $G$, $D$ and $R$. The residual blocks (ResBlocks) we used are detailed in Fig. \ref{fig:2_resblock}. 


The network $\varphi$ first embeds each character of the sequence $\boldsymbol{s}$ in $\mathbb{R}^{128}$, then encodes it with a four-layer bidirectional LSTM \cite{Hochreiter,Gers2000} recurrent network (with a hidden state of size 128). $\varphi(\boldsymbol{s})$ is the output of the last bidirectional LSTM layer.

The network $G$ is derived from \cite{Brock18}. The input noise, of dimension 128, is split into eight equal-sized chunks. The first one is passed to a fully connected layer of dimension 1024, whose output is reshaped to $256 \times 1 \times 4$ (with the convention $\mathrm{depth} \times \mathrm{height} \times \mathrm{width}$). Each of the seven remaining chunks is concatenated with the embedding $\varphi(\boldsymbol{s})$, and fed to an up-sampling ResBlock through Conditional Batch Normalization (CBN) \cite{de2017modulating} layers (see Fig. \ref{fig:2_resblock}). The consecutive ResBlocks have the following number of filters: 256, 128, 128, 64, 32, 16, 16. A self-attention layer \cite{Zhang2018} is used between the fourth and the fifth ResBlocks. We add a final convolutional layer and a tanh activation in order to obtain a $1 \times 128 \times 512$ image.

The network $D$ is made up of seven down-sampling ResBlocks (with the following number of filters: 16, 16, 32, 64, 128, 128, 256), a self-attention layer between the third and the fourth ResBlocks, and a normal ResBlock (with 256 filters). We then sum the output along horizontal and vertical dimensions and project it on $\mathbb{R}$.

The auxiliary network $R$ is a Gated Convolutional Network, introduced in \cite{Bluche2017} (we used the ``big architecture''). This network consists in an encoder of five convolutional layers, with Tanh activations and convolutional gates, followed by a max pooling on the vertical dimension and a decoder made up of two stacked bidirectional LSTM layers. 


\subsection{Optimization settings} 
\label{subsec:optim}

We used spectral normalization \cite{MiyatoSpectral18} in $G$ and $D$, following recent works \cite{Zhang2018,Brock18,MiyatoSpectral18} that found that it stabilizes the training. We optimized our system with the Adam algorithm \cite{kingma2014adam} (for all networks: $lr = 2\times10^{-4}, \beta_1 = 0, \beta_2 = 0.999$) and we used gradient clipping in $D$ and $R$. We trained our model for several hundred thousand iterations with mini-batches of 64 images of the same type, either real or generated.   

While $D$ processes one real batch and one generated batch per training step, $R$ is trained with real data only, to prevent it from learning how to recognize generated (and potentially false) images of text. To train the networks $G$ and $\varphi$, we first produce a batch of ``fake'' images $\boldsymbol{x_{\mathrm{fake}}} := G(\boldsymbol{z}, \varphi(\boldsymbol{s}))$, and then pass it through $D$ and $R$. $(G, \varphi)$ learn from the gradients $\boldsymbol{\nabla_D}  := -\frac{\partial D(\boldsymbol{x_{\mathrm{fake}}})}{\partial \boldsymbol{x_{\mathrm{fake}}}}$ and $\boldsymbol{\nabla_R} := \frac{\partial\mathrm{CTC}(\boldsymbol{s}, R(\boldsymbol{x_{\mathrm{fake}}}))}{\partial \boldsymbol{x_{\mathrm{fake}}}}$ coming from these two networks. Since $R$ and $D$ have different architectures and losses, the norms of $\boldsymbol{\nabla_D}$ and $\boldsymbol{\nabla_R}$ can differ by several orders of magnitudes (we observed that $||\boldsymbol{\nabla_R}||_2$ is typically $10^2$ to $10^3$ times greater than $||\boldsymbol{\nabla_D}||_2$). To have $(G, \varphi)$ learn from both $D$ and $R$, we found it useful to balance the two gradients before propagating them to $G$. Therefore, we apply the following affine transformation to $\boldsymbol{\nabla_R}$:

\[
\boldsymbol{\nabla_R} \leftarrow \alpha \times \big( \frac{\sigma_D}{\sigma_R} (\boldsymbol{\nabla_R} - \mu_R ) + \mu_D \big)
\]

\noindent With $\mu_\bullet$ and $\sigma_\bullet$ being the mean and the standard deviation of $\boldsymbol{\nabla_\bullet}, \bullet \in \{D, R\}$. $\alpha$ controls the relative importance of $R$ with respect to $D$ and is set to $1$ in our model. The concrete impact of this transformation is discussed in Section \ref{subsubsec:gb}.  



\section{Results}
\label{sec:results}

\subsection{Experimental setup}
\label{subsec:setup}

\begin{figure}[h]
    \centering
    \includegraphics[width=\columnwidth]{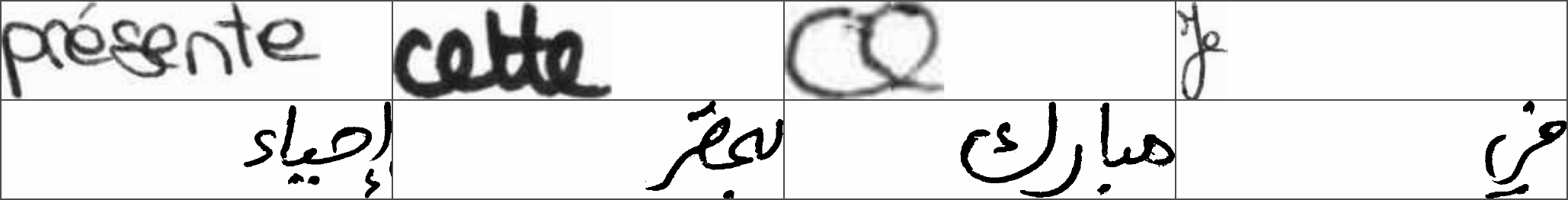}
    \caption{Images after preprocessing. First line: RIMES. Second line: OpenHaRT.}
    \label{fig:3_data}
\end{figure}

In our experiments, we use $128 \times 512$ images of handwritten words obtained with the following preprocessing: we isometrically resize the images to a height of 128 pixels, then remove the images of width greater than 512 pixels and finally, pad them with white to reach a width of 512 pixels for all the images (right-padding for French, left-padding for Arabic). Table \ref{tab:3_data} summarizes the meaningful characteristics of the two datasets we work with, namely the RIMES \cite{grosicki2009icdar} and the OpenHaRT \cite{openhart2010} datasets, while Fig. \ref{fig:3_data} shows some images from these two datasets. 

\begin{table}[h!]
    \centering
    \caption{Characteristics of the subsets of RIMES and OpenHaRT.}
    \label{tab:3_data}
    \begin{tabular}{|c|c|c|c|c|c|}
        \hline
        Dataset     & Language  & Images    & Words     & Characters    \\
        \hline
        RIMES       & French    & 129414    & 6780      & 86            \\
        OpenHaRT    & Arabic    & 710892   & 65575      & 77            \\
        \hline
    \end{tabular}   
\end{table}

To reflect the distribution found in natural language, the words to be generated are sampled from a large list of words (French Wikipedia for French, OpenHaRT for Arabic). For the text recognition experiments on the RIMES dataset (Section \ref{subsec:reco}), we use a separate validation dataset of 6060 images.


We evaluate the performance with Fr\'{e}chet Inception Distance (FID) \cite{FID2017} and Geometry Score (GS) \cite{Khrulkov2018}. FID is widely used and gives a distance between real and generated data. GS compares the topology of the underlying real and generated manifolds and provides a way to measure the mode collapse. For these two indicators, lower is better. In general, we observed that FID correlates with visual impression better than GS. For each experiment, we computed FID (with 25k real and 25k generated images) and GS (with 5k real and 5k generated images, 100 repetitions and default settings for the other parameters) every 10000 iterations and trained the system with different random seeds. We then chose independently the best FID and the best GS among the different runs. To verify the textual content, we relied on visual inspection. To measure the impact of data augmentation on the text recognition performance, we used Levenshtein distance at the character level (Edit Distance) and Word Error Rate. 

%

\subsection{Ablation study}
\label{subsec:ablation}

For all the experiments in this section, we used the RIMES database described in Section \ref{subsec:setup}. 


\subsubsection{Gradient balancing}
\label{subsubsec:gb}

When training the networks $(G, \varphi)$, the norms of the gradients coming from $D$ and $R$ may differ by several orders of magnitudes. As mentioned in Section \ref{sec:model}, we found it useful to balance these two gradients. Table \ref{tab:gb} reports FID, GS and a generated image for different gradient balancing settings. 

\begin{table}[h!]
    \centering
    \caption{FID, GS and a generated image of the word ``r\'{e}parer'', for four settings: no gradient balancing, $\alpha=0.1$, $\alpha=1$ (our model) and $\alpha=10$.}
    \label{tab:gb}
    \begin{tabular}{|c||c|c|c|}
        \hline
        $\alpha$    & FID       & GS                    & Images            \\
        \hline
        None        & 141.35    & $2.44 \times 10^{-3}$ &   \begin{minipage}{3cm}
                                                                \centering
                                                                \includegraphics[width=\linewidth]{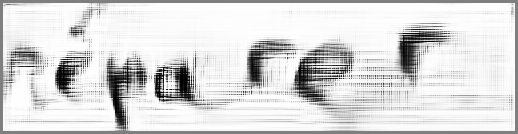}
                                                            \end{minipage}  \\
                                                            
        0.1         & 72.93     & $4.23 \times 10^{-2}$ &   \begin{minipage}{3cm}
                                                                \centering
                                                                \includegraphics[width=\linewidth]{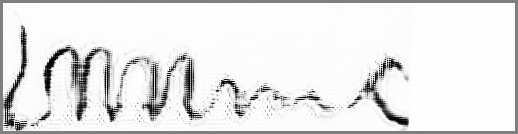}
                                                            \end{minipage}  \\

        10          & 222.47    & $2.92 \times 10^{-3}$ &   \begin{minipage}{3cm}
                                                                \centering
                                                                \includegraphics[width=\linewidth]{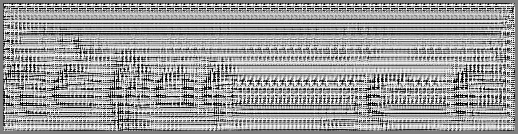}
                                                            \end{minipage}  \\

        \textbf{1}  & \textbf{23.94}    & $\mathbf{8.58 \times 10^{-4}}$  &   \begin{minipage}{3cm}
                                                                \centering
                                                                \includegraphics[width=\linewidth]{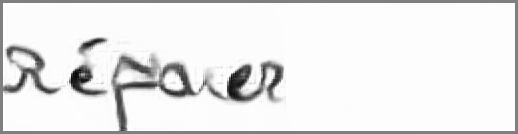}
                                                            \end{minipage}  \\
        \hline
    \end{tabular}   
\end{table}

Without gradient balancing, we observed that $||\boldsymbol{\nabla_R}||_2$ was typically $10^2$ to $10^3$ times greater than $||\boldsymbol{\nabla_D}||_2$, meaning that the learning signal for $(G, \varphi)$ is biased toward satisfying $R$. As a result, the word ``r\'{e}parer'' is clearly readable, but the FID is high (141.35) and the generated image is not realistic (the background is noisy, the letters are too far apart).   

With $\alpha=0.1$, $||\boldsymbol{\nabla_R}||_2$ is much smaller than $||\boldsymbol{\nabla_D}||_2$, meaning that $G$ and $\varphi$ take little account of the auxiliary recognition task. As illustrated by the second image in Table \ref{tab:gb}, we lose control of the textual content of the generated image. FID is better than before, but still high (72.93). In a way, the generated image is quite realistic, since the background is whiter and the writing more cursive.

On the contrary, when setting $\alpha$ to 10, $G$ and $\varphi$ mostly learn from the feedback of $R$ and the generation is thus successfully conditioned on the textual content. In fact, we can distinguish the letters of ``r\'{e}parer'' in the third generated image in Table \ref{tab:gb}. However, as we are focusing on optimizing the generation process to have a minimal CTC cost, we observe strong visual artifacts that remind of the one obtained by Deep Dream generators \cite{mordvintsev2015inceptionism}. FID is much higher (222.47) and the resulting images are very noisy, as demonstrated by the third image in Table \ref{tab:gb}. 

The best compromise corresponds to $\alpha=1$. We obtain the best FID of 23.94 and GS of \num{8.58e-4}, while the generated image is both readable and realistic. For all other experiments, we set $\alpha$ to 1. 

\subsubsection{Adversarial loss}

Using the network architecture described in Section \ref{sec:model}, we test three different adversarial training procedures: the ``vanilla'' GAN \cite{NIPS2014_5423} (GAN), the Least Squares GAN \cite{mao2016least} (LSGAN) and the Geometric GAN \cite{Lim2017, Zhang2018, Brock18}, used in our model. FID and GS are reported in Table \ref{tab:3_losses}. 

\begin{table}[h]
    \centering
    \caption{FID and GS for different adversarial losses.}
    \label{tab:3_losses}
    \begin{tabular}{|c||c|c|c|c|}
        \hline
        Adversarial Loss        & FID           & GS                \\
        \hline
        GAN                     & 36.32         & $5.29 \times 10^{-3}$          \\
        LSGAN                   & 116.09        & $3.78 \times 10^{-3}$          \\
        \textbf{Geometric GAN}  & \textbf{23.94}& $\mathbf{8.58 \times 10^{-4}}$ \\
        \hline
    \end{tabular}
\end{table}

As shown in Table \ref{tab:3_losses}, Geometric GAN leads to the best performance in terms of FID and GS. LSGAN fails to produce text-like outputs in three out of five trials. The low FID for vanilla GAN indicates that it produces realistic images. The high GS in Table \ref{tab:3_losses} shows that both GAN and LSGAN suffer from a style collapse, and we observed that the textual content was not controlled. The trends given by FID and GS have been successfully confirmed by visual inspection of the generated samples.

\subsubsection{Self-attention}
We use a self-attention layer \cite{Zhang2018}, in both the generator and the discriminator, as it may help to keep coherence across the full image. We trained our model with and without this module to measure its impact. 

\begin{table}[h]
    \centering
    \caption{Impact of self-attention.}
    \label{tab:3_sa}
    \begin{tabular}{|c||c|c|c|c|}
        \hline
                                        & FID               & GS                            \\
        \hline
        Without self-attention          & 67.86             & $4.51 \times 10^{-3}$         \\
        \textbf{With self-attention}    & \textbf{23.94}    & $\mathbf{8.58 \times 10^{-4}}$ \\
        \hline
    \end{tabular}
\end{table}

Without self-attention, we still obtain realistic samples with correct textual content, but using self-attention improves performance both in terms of FID and GS, as shown in Table \ref{tab:3_sa}.

\subsubsection{Conditional Batch Normalization}

As described in Section \ref{sec:model}, $G$ is provided a noise chunk and $\varphi(s)$ through each CBN layer. Another reasonable option, closer to \cite{Odena2016}, is to concatenate the whole noise $\boldsymbol{z}$ with $\varphi(\boldsymbol{s})$, and feed it to the first linear layer of $G$ (in this scenario, CBN is replaced with standard Batch Normalization). Table \ref{tab:concat} reports FID and GS for these two solutions. 

\begin{table}[h]
    \centering
    \caption{Generator input via the first linear layer or via CBN layers.}
    \label{tab:concat}
    \begin{tabular}{|c||c|c|}
        \hline
                    & FID   & GS    \\
        \hline
        First linear layer  & 42.23             & $1.81 \times 10^{-3}$         \\
        \textbf{CBN layers} & \textbf{23.94}    & $\mathbf{8.58 \times 10^{-4}}$\\
        \hline
    \end{tabular}
\end{table}

FID and GS in Table \ref{tab:concat} indicates that feeding the generator inputs through CBN layers improves realism and reduces mode collapse. The visual inspection of the generated samples confirmed these trends and showed that the other solution prevents from correctly conditioning on the textual content. \\

%
%

\subsection{Generation of handwritten text images}

We trained the model detailed in Section \ref{sec:model} on the two datasets described in Section \ref{subsec:setup}, RIMES and OpenHaRT. 

\begin{figure}[h]
    \centering
    \includegraphics[width=0.8\columnwidth]{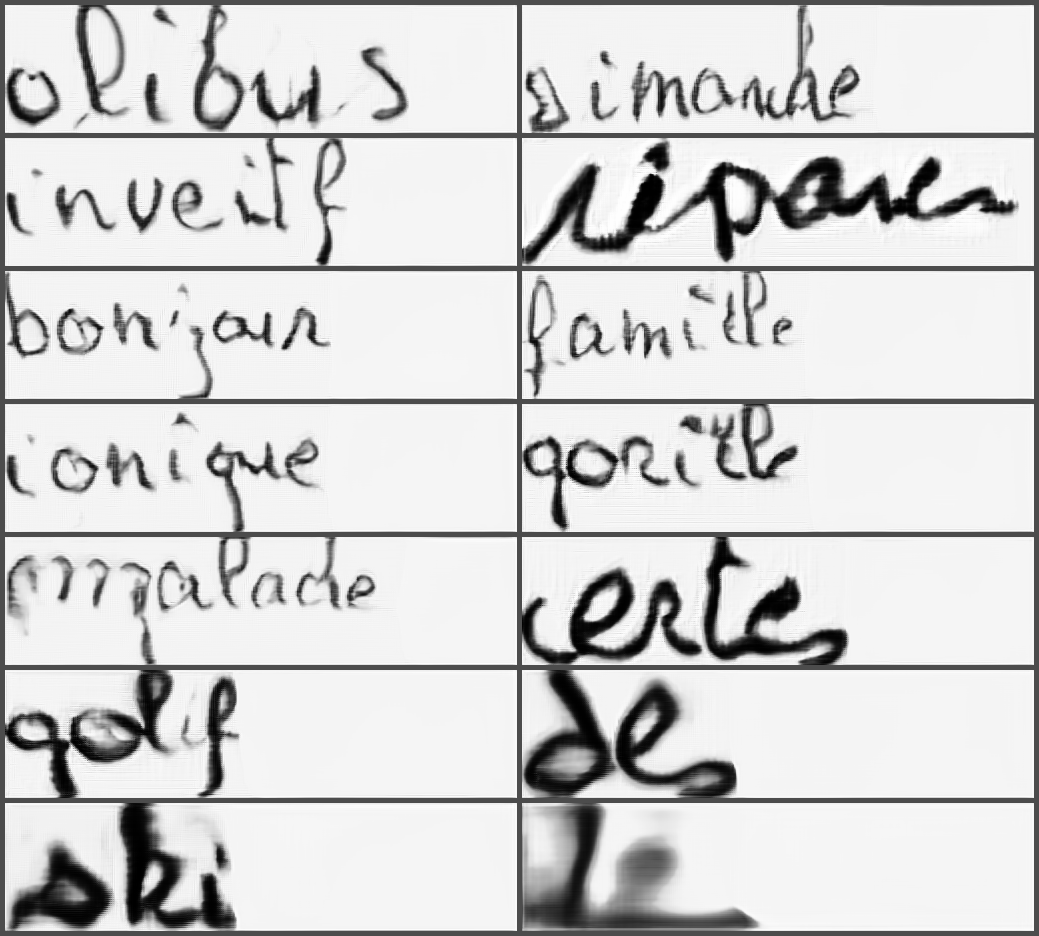}
    \caption{Images generated with our system trained on RIMES. Targets: \textit{olibrius}, \textit{Dimanche}, \textit{inventif}, \textit{r\'{e}parer}, \textit{bonjour}, \textit{famille}, \textit{ionique}, \textit{gorille}, \textit{malade}, \textit{certes}, \textit{golf}, \textit{des}, \textit{ski}, \textit{le}.}
    \label{fig:3_french}
\end{figure}

\begin{figure}[h]
    \centering
    \includegraphics[width=0.8\columnwidth]{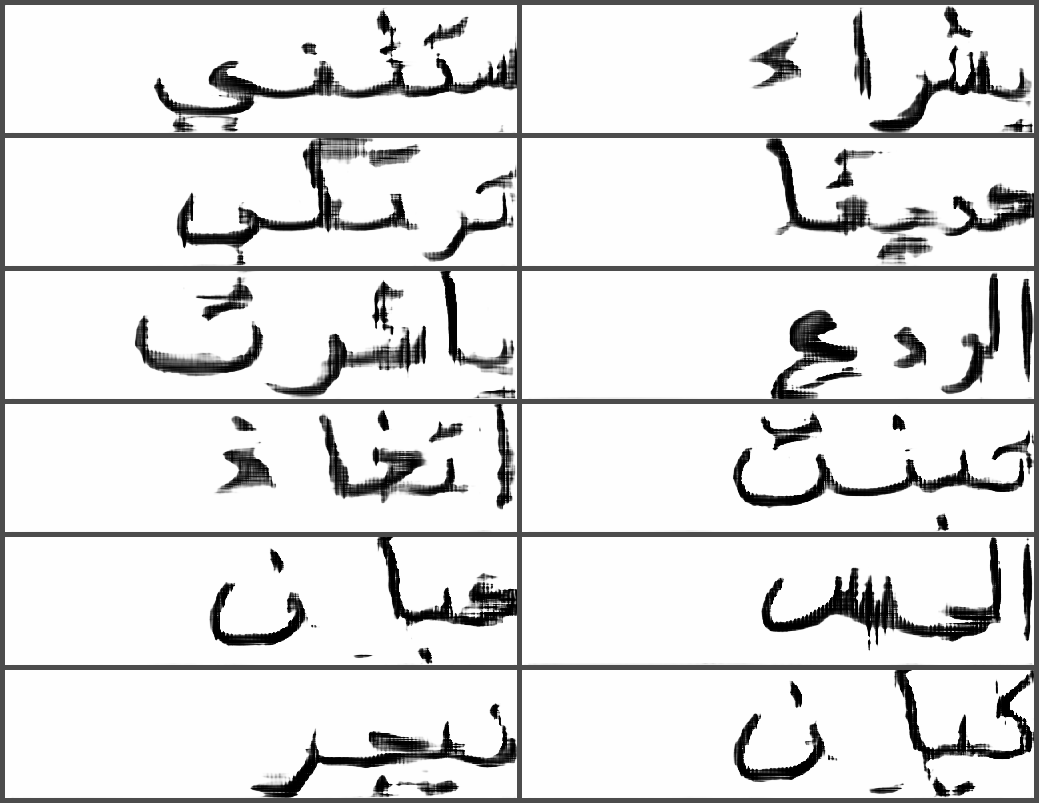}
    \caption{Images generated with our system trained on OpenHaRT. Targets: \novocalize \<st_tny>, \<b^srA'>, \<trtkb>, \<.hdy_tA>, \<bA^srt>, \<Alrd`>, \<AtxA_d>, \<tbnt>, \<.hbAn>, \<Al.hs>, \<ny^gr>, \<kyAn>.  }
    \label{fig:3_arabic}
\end{figure}

Fig. \ref{fig:3_french} and Fig. \ref{fig:3_arabic} display some randomly generated (not cherry-picked) samples in French and Arabic respectively. For these two languages, we observe that our model is able to produce images of cursive handwriting, successfully conditioned on variable-length words (even if some words remain barely readable, e.g. \textit{le} and \textit{olibrius} in Fig. \ref{fig:3_french}). The typography of the individual characters is varied, but we can detect a slight collapse of writing style among the images. For French, as we trained the generator to produce words from all Wikipedia, we are able to successfully synthesize words that are not present in the training dataset. In Fig. \ref{fig:3_french} for instance, the words \textit{olibrius}, \textit{inventif}, \textit{ionique}, \textit{gorille} and \textit{ski} are not in RIMES, while \textit{Dimanche}, \textit{bonjour}, \textit{malade} and \textit{golf} appear in the corpus but with a different case.

\subsection{Data augmentation for handwritten text recognition}
\label{subsec:reco}

We aim at evaluating the benefits of generated data to train a model for handwritten text recognition. To this end, we trained from scratch a Gated Convolutional Network \cite{Bluche2017} (identical to the network $R$ described in Section \ref{subsec:archi}) with the CTC loss, RMSprop optimizer \cite{rmsprop} and a learning rate of $10^{-4}$. We used the validation data described in \ref{subsec:setup} for early stopping.

\begin{table}[h]
    \centering
    \caption{Extending the RIMES dataset with 100k generated images. Impact on the text recognition performance in terms of Edit Distance (ED) and Word Error Rate (WER) on the validation set.}
    \label{tab:data}
    \begin{tabular}{|l|c|c|}
        \hline
        Data            & ED    & WER   \\
        \hline
        RIMES only      & 4.34  & 12.1  \\
        RIMES + 100k    & 4.03  & 11.9  \\
        \hline
    \end{tabular}
\end{table}

Table \ref{tab:data} shows that extending the RIMES dataset with data generated with our adversarial model brings a slight improvement in terms of Edit Distance and Word Error Rate. Note that using only GAN-made synthetic images for training the text recognition model does not yield competitive results.


\section{Conclusion}

We presented an adversarial model to produce synthetic images of handwritten word images, conditioned on the sequence of characters to render. Beyond the classical use of a generator and a discriminator to create plausible images, we employ recurrent layers to embed the word to condition on, and add an auxiliary recognition network in order to generate an image with legible text. Another crucial component of our model lies in balancing the gradients coming from the discriminator and from the recognizer when training the generator. 

We obtained realistic word images in both French and Arabic. 
Our experiments showed a slight reduction in error rate for the French model trained on combined data.

An immediate continuation of our experiments would be to train the described model on more challenging datasets, with textured background for instance. Furthermore, deeper investigation to reduce the observed phenomenon of style collapse would be a significant improvement. Another important line of work is to extend this system to the generation of line images of varying size.



\bibliographystyle{IEEEtran}
\bibliography{IEEEabrv,articleGAN}


%
%

\end{document}